\documentstyle{elsart1p}
\newtheorem{theorem}{Theorem}

\newtheorem{corollary}{Corollary}

\begin{document}

\begin{frontmatter}



\title{A game-theoretic version of Oakes' example for randomized forecasting}


\author{Vladimir V. V'yugin\thanksref{label1}}
\thanks[label1]{This research was partially supported by Russian foundation
for fundamental research: 06-01-00122-a.
}

\address{Institute for Information Transmission Problems,
Russian Academy of Sciences,
Bol'shoi Karetnyi per. 19, Moscow GSP-4, 127994, Russia.
e-mail vyugin@iitp.ru}

\begin{abstract}
Using the game-theoretic framework for probability, Vovk and
Shafer~\cite{ShV2005}
have shown that it is always possible, using randomization,
to make sequential probability forecasts that pass any countable
set of well-behaved statistical tests. This result generalizes work by
other authors, who consider only tests of calbration.

We complement this result with a lower bound.
We show that Vovk and Shafer's result is valid only when the forecasts are
computed with unrestrictedly increasing degree of accuracy.

When some level of discreteness is fixed, we present a game-theoretic generalization of
Oakes' example for randomized forecasting that is a test failing
any given method of deferministic forecasting; originally, this example was
presented for deterministic calibration.
\end{abstract}

\end{frontmatter}

\maketitle

\section{Introduction}\label{intr-1}

Using the game-theoretic framework for probability~\cite{ShV2001},
Vovk and Shafer have shown in~\cite{ShV2005}
that it is always possible, using randomization,
to make sequential probability forecasts that pass any countable set
of well-behaved
statistical tests. This result generalizes work by other authors,
among them are Foster and Vohra~\cite{FoV98}, Kakade and Foster~\cite{KaF2004},
Lehrer~\cite{Lehr97}, Sandrony et al.~\cite{sansmovo03}, who consider
only tests of calibration.

We complement this result with a lower bound.
We show that Vovk and Shafer's result is valid only when the forecasts are
computed with unrestrictedly increasing degree of accuracy. When some level
of discreteness is fixed, we present a game-theoretic version of
Oakes' example for randomized forecasting that is a test failing
any given method of deterministic forecasting; originally, this example was
presented for deterministic calibration. To formulate this example, we
use the forecasting game presented by Vovk and Shafer~\cite{ShV2005},
namely Binary Forecasting Game II.

We discuss details of the randomized forecasting algorithms in
Section~\ref{wheat-1}.

The Shafer and Vovk's~\cite{ShV2001} game-theoretic framework
is considered in Section~\ref{game-1}. We present in this section the original
Vovk and Shafer's~\cite{ShV2005} result on universal randomized forecasting
and prove our result which gives the limits for such forecasting -
a game-theoretic version of the Oakes' example for randomized
forecasting.

\section{Background}\label{wheat-1}

The research discussed in this paper was started from a notion of calibration,
originated by Dawid~\cite{Daw82, Daw85}. A test of calibration
checks whether the observed empirical frequencies of state occurrences
converge to their forecaster probabilities.

Let $I(p)$ denote the indicator function of a
subinterval $I\subseteq [0,1]$, i.e., $I(p)=1$ if $p\in I$, and $I(p)=0$,
otherwise.
An infinite sequence of forecasts $p_1,p_2,\dots$ is {\it well-calibrated}
for an infinite sequence of outcomes
$\omega_1\omega_2\dots$ if for the characteristic function $I(p)$
of any subinterval of $[0,1]$ {\it the calibration error} tends to zero, i.e.,
\begin{eqnarray}\label{call-1}
\frac{\sum_{i=1}^{n}I(p_i)(\omega_i-p_i)}
{\sum_{i=1}^{n}I(p_i)}\to 0
\end{eqnarray}
as the denominator of the relation (\ref{call-1}) tends to infinity.
Here, $I(p_i)$ determines some
``selection rule'' which defines indices $i$ where we compute
the deviation between forecasts $p_i$ and outcomes $\omega_i$.

The main problem of sequential forecasting is to define a universal
forecasting algorithm which computes forecasts $p_n$ given past observations
$\omega_1,\dots ,\omega_{n-1}$ for each $n$. This universal prediction
algorithm should be well-calibrated for each infinite sequence of outcomes.
Oakes~\cite{Oak85} proposed arguments (see Dawid~\cite{Daw85a}
for a different proof) that no such algorithm can be
well-calibrated for all possible sequences: any forecasting
algorithm cannot be calibrated for the sequence
$\omega=\omega_1\omega_2\dots$, where
\[
\omega_i=
  \left\{
    \begin{array}{l}
      1 \mbox{ if } p_i<0.5
    \\
      0 \mbox{ otherwise }
    \end{array}
  \right.
\]
and $p_i$ are forecasts computed by the algorithm given
$\omega_1,\dots ,\omega_{i-1}$, $i=1,2,\dots$.
The corresponding intervals are $I_0=[0,0.5)$ and $I_1=[0.5,1]$.
It is easy to see that the condition (\ref{call-1}) of calibration fails
for this $\omega$, where $I=I_0$ or $I=I_1$.

Foster and Vohra~\cite{FoV98} show that calibration is almost surely
guaranteed with a randomizing forecasting rule, i.e., where the forecasts
are chosen using internal randomization.
Kakade and Foster~\cite{KaF2004} noticed that some calibration results
require very little randomization. They defined ``an almost deterministic''
{\it randomized rounding} universal forecasting algorithm $f$:
for any sequence of outcomes $\omega_1\omega_2\dots$, an observer can only
randomly round the deterministic forecast up to $\Delta$ in order to calibrate
with the internal probability $1$
\begin{eqnarray}\label{call-1g}
\left|\frac{1}{n}\sum_{i=1}^{n}I(p_i)(\omega_i-p_i)\right|\le\Delta,
\end{eqnarray}
where $\Delta$ is the calibration error, $I(p)$ is the indicator function of
an arbitrary subinterval of $[0,1]$.

This approach was further developed by, among others, Lehrer~\cite{Lehr97},
Sandrony et al.~\cite{sansmovo03}.
These papers were only concerned with asymptotic calibration.
Non-asymptotic version of randomized forecasting was proposed by
Vovk and Shafer~\cite{ShV2005} and by Vovk et al.~\cite{Vov2005}.
They based on the game-theoretic framework of Shafer and Vovk~\cite{ShV2001}.

Let ${\cal P}\{0,1\}$ be the set of all measures on
the two-element set $\{0,1\}$. Any measure from ${\cal P}\{0,1\}$
is represented by a number $p\in [0,1]$ - the probability of $\{1\}$.
We consider also the set of all measures on ${\cal P}\{0,1\}$.
Let ${\cal P}[0,1]$ be the set of all probability measures on the
unit interval $[0,1]$ supplied with the standard Borel $\sigma$-field
$\cal F$.

Randomizing forecasting is defined as follows.
For each $n$, a forecaster given a binary sequence of past outcomes
$\omega_1\dots\omega_{n-1}$ (and a sequence of past forecasts
$p_1,\dots ,p_{n-1}$) outputs a probability distribution
$P_n\in{\cal P}[0,1]$. The forecasts $p_n$ of the
the future event $\omega_n=1$ are distributed according to this
probability distribution.

Assume for each $n$, the probability distribution
$P_n$ is concentrated on a finite subset $D_n$ of $[0,1]$, say,
$D_n=\{p_{n,1},\dots , p_{n,m_n}\}$. The number
$\Delta=\liminf\limits_{n\to\infty}\Delta_n$, where
$$
\Delta_n=\inf\{|p_{n,i}-p_{n,j}|:i\not =j\},
$$
is called {\it the level of discreteness} of the corresponding
forecasting scheme on the sequence $\omega_1\omega_2\dots$.

In general case $D_n$ is a predictable random variable,
i.e., measurable with respect to the $\sigma$-field ${\cal F}^{n-1}$,
depending on $\omega_1\dots\omega_{n-1}$.

A typical example is the uniform rounding: for each $n$,
rational points $p_{n,i}$ divide the unit interval into equal
parts of size $0<\Delta<1$ and $P_n$ is concentrated on these points.
In this case the level of discreteness on arbitrary sequence
$\omega_1\omega_2\dots$ equals $\Delta$.

\section{Game-theoretic randomized forecasting}\label{game-1}

Shafer and Vovk~\cite{ShV2001} proposed a game-theoretic framework for
probability theory. In Vovk and Shafer~\cite{ShV2005} they used this
framework to demonstrate the possibility of good probability forecasting
in a general setting. This result generalizes the previous work of
many authors.

Vovk and Shafer presented a game between Reality, Forecaster and Skeptic.
In this game, Forecaster faces Skeptic whose strategy is revealed in
advance, and he is allowed to use a degree of randomization to conceal each
of his probability forecasts until the corresponding outcome has been
announced. Their main result says that Forecaster can keep
Skeptic from becoming infinitely reach. Intuitively, this means that
the outcomes determined by Reality look random with respect to
probability forecasts. This result is a consequence of the von Neumann's
minimax theorem.

Vovk and Shafer~\cite{ShV2005} consider a perfect-information game of
randomized forecasting - {\it Binary Forecasting Game II}
between three players - Forecaster, Skeptic, Reality, Random Number Generator
described by the following {\it protocol}:

Let ${\cal K}_0=1$ and ${\cal F}_0=1$.\\
FOR $n=1,2,\dots$\\
Skeptic announces $S_n:[0,1]\to\cal R$.\\
Forecaster announces a probability distribution $P_n\in{\cal P}[0,1]$.\\
Reality announces $\omega_n\in\{0,1\}$.\\
Forecaster announces $f_n:[0,1]\to\cal R$ such that $\int f_n(p)P_n(dp)\le 0$.\\
Random Number Generator announces $p_n\in [0,1]$.\\
Skeptic updates his capital
${\cal K}_n={\cal K}_{n-1}+S_n(p_n)(\omega_n-p_n)$.\\
Forecaster updates his capital ${\cal F}_n={\cal F}_{n-1}+f_n(p_n)$.
\\
ENDFOR

{\it Restriction on Skeptic:} Skeptic
must choose the $S_n$ so that his capital ${\cal K}_n$
is nonnegative for all $n$ no matter how the other players move.

{\it Restriction on Forecaster:} Forecaster
must choose the $P_n$ and $f_n$ so that his capital ${\cal F}_n$
is nonnegative for all $n$ no matter how the other players move.


Vovk and Shafer~\cite{ShV2005} showed that Forecaster has a winning
strategy in the Forecasting Game II, where
Forecaster wins if either (i) his capital ${\cal F}_n$ is unbounded
or (ii) Skeptic's capital ${\cal K}_n$ stays bounded; otherwise the
other players win.
\begin{theorem}\label{rand-1}
Forecaster has a winning strategy in Binary Forecasting Game II.
\end{theorem}
{\it Sketch of the proof}. For completeness of the presentation,
we reproduce the proof from~\cite{ShV2005}.
The proof is based on von Neumann's minimax theorem.

At first, at any round $n$ of Binary Forecasting Game II, a simple auxiliary
game between Realty and Forecaster is considered:
Forecaster chooses $p_n\in [0,1]$, Realty chooses $\omega_n\in\{0,1\}$.
Forecaster losses (and Realty gains) $S(p_n)(\omega_n-p_n)$.

For any mixed strategy of Realty $Q_n\in {\cal P}\{0,1\}$, let Forecaster's
strategy be $p_n=Q\{1\}$. So, the Realty's expected gain
is $S(p_n)(1-Q\{1\})Q\{1\}+S(p_n)(0-Q\{1\})(1-Q\{1\})=0$.

In order to apply von Neumann's minimax theorem, which requires that move space
be finite, we replace Forecaster move space $[0,1]$ with a
finite subset of $[0,1]$ dense enough that the value of the game
is smaller than some arbitrary small positive number $\Delta$
(depending on $n$). This is possible,
since $|S_n(p)|\le{\cal K}_{n-1}\le 2^{n-1}$. 
\footnote
{
Skeptic must choose $S_n(p)$ such that ${\cal K}_n\ge 0$ for all $n$
no matter the other players move.
}
The minimax theorem asserts that Forecaster
has a mixed strategy $P\in {\cal P}[0,1]$ such that
\begin{eqnarray}\label{int-1}
\int S_n(p)(\omega_n-p)P(dp)\le\Delta
\end{eqnarray}
for both $\omega_n=0$ and $\omega_n=1$.

Let $E_\Delta$ be the subset of ${\cal P}[0,1]$ consisting
all probability measures $P$ satisfying (\ref{int-1}) for
$\omega_n=0$ and $\omega_n=1$. Endowed with the weak
topology, ${\cal P}[0,1]$ is compact. Since each $E_\Delta$ is closed,
${\cap E}_{\Delta_i}\not =\emptyset$, where $\Delta_i$, $i=1,2,\dots$,
is some decreasing to $0$ sequence of real numbers. So there exists
$P_n\in{\cal P}[0,1]$ such that
\begin{eqnarray*}\label{int-2}
\int S_n(p)(\omega_n-p)P_n(dp)\le 0
\end{eqnarray*}
for both $\omega_n=0$ and $\omega_n=1$.

In Binary Forecasting Game II, consider the strategy for Forecaster that
uses at any round $n$ the probability distribution $P_n$ just defined
and uses as his second move the function $f_n$ defined
$f_n(p)=S_n(p)(\omega_n-p)$. Then ${\cal F}_n={\cal K}_n$ for all $n$.
So either Skeptic's capital will stay bounded or Forecaster's
capital will be unbounded.
$\triangle$

Vovk et al.~\cite{Vov2005} (see also~\cite{ShV2005}) also showed that
Skeptic can present a strategy
$S_n(p)$ such that the winning strategy of Forecaster existing
by Theorem~\ref{rand-1} announces forecasts $p_1,p_2,\dots$ which are
well-calibrated for an arbitrary sequence $\omega_1\omega_2\dots$ of outcomes.

In that follows we consider some version of the Oakes'
example in the game-theoretic framework. A different version
of this result is given in~\cite{Vyu2007}.

We consider some modification of Binary Forecasting Game II
in which Skeptic (but not Forecaster) announces $f_n:[0,1]\to\cal R$.
This means that Skeptic defines the test of randomness he needs.

Also, at each step $n$, Skeptic divide his capital into two accounts:
${\cal K}_n={\cal Q}_n+{\cal F}_n$; he uses the capital ${\cal F}_n$
to force Random Number Generator to generate random numbers which pass 
the test $f_n$.

Let ${\cal K}_0=2$.\\
FOR $n=1,2,\dots$\\                              
Skeptic announces $S_n:[0,1]\to\cal R$.\\
Forecaster announces a probability distribution $P_n\in{\cal P}[0,1]$.\\
Reality announces $\omega_n\in\{0,1\}$.\\
Skeptic announces $f_n:[0,1]\to\cal R$ such that $\int f_n(p)P_n(dp)\le 0$.\\
Random Number Generator announces $p_n\in [0,1]$.\\
Skeptic updates his capital
${\cal K}_n={\cal K}_{n-1}+S_n(p_n)(\omega_n-p_n)+f_n(p_n)$.\\
ENDFOR

We divide the Skeptic's capital into two parts:\\
${\cal K}_n={\cal Q}_n+{\cal F}_n$ for all $n$, where\\
${\cal Q}_0=1$ and ${\cal F}_0=1$.\\
${\cal Q}_n={\cal Q}_{n-1}+S_n(p_n)(\omega_n-p_n)$ and\\
${\cal F}_n={\cal F}_{n-1}+f_n(p_n)$.

{\it Restriction on Skeptic:} Skeptic
must choose the $S_n$ and $f_n$ so that his capital ${\cal K}_n$
is nonnegative for all $n$ no matter how the other players move.

Actually, Skeptic will choose the $S_n$ and $f_n$ so that both of his 
capitals ${\cal Q}_n$ and 
${\cal F}_n$ are nonnegative for all $n$ no matter how the other players move.

We prove that when Forecaster uses finite subsets of $[0,1]$ for randomization
Realty and Skeptic can defeat Forecaster (and Random Number Generator) 
in this forecasting game, where Realty and Skeptic win if Skeptic's capital 
${\cal K}_n$ is unbounded; otherwise Forecaster and Random Number Generator win.

\begin{theorem}\label{game-oakes}
Assume Forecaster's uses a randomized strategy with
a positive level of discreteness on each infinite sequence $\omega$.
Then Realty and Skeptic win in the modified Binary Forecasting Game II.
\end{theorem}
{\it Proof}.
Define a strategy for Realty: at any step $n$ Realty announces
an outcome
\[
  \omega_n=
  \left\{
    \begin{array}{l}
      0 \mbox{ if } P_n((0.5,1])>0.5
    \\
      1 \mbox{ otherwise. }
    \end{array}
  \right.
\]

We follow Shafer and Vovk's~\cite{ShV2001} method of defining
the defensive strategy for Skeptic.

Let $\epsilon_k=2^{-k}$, $k=1,2,\dots$. We define recursively by $n$: 
${\cal Q}^{s,k}_0=1$, $S^{s,k}_0(p)=0$, $s=1,2$, and for $n\ge 1$  
\begin{eqnarray}
S^{1,k}_n(p)=-\epsilon_k {\cal Q}^{1,k}_{n-1}\xi(p>0.5),
\label{strat-1}
\\
S^{2,k}_n(p)=\epsilon_k {\cal Q}^{2,k}_{n-1}\xi(p\le 0.5),
\label{strat-2}
\end{eqnarray}
where $\xi(true)=1$, $\xi(false)=0$, and for $n\ge 1$
\begin{eqnarray}
{\cal Q}_{n}^{1,k}={\cal Q}_{n-1}^{1,k}+S^{1,k}_{n}(p_{n})
(\omega_{n}-p_{n})),
\label{cap-1t}
\\
{\cal Q}_{n}^{2,k}={\cal Q}_{n-1}^{2,k}+S^{2,k}_{n}(p_{n})
(\omega_{n}-p_{n})).
\label{cap-2t}
\end{eqnarray}
We combine $S^{1,k}_n(p)$ and $S^{2,k}_n(p)$ in the Skeptic's strategy 
$S_n(p)=\frac{1}{2}(S^1_n(p)+S^2_n(p))$,
where 
$$
S^1_n(p)=\sum\limits_{k=1}^\infty\epsilon_k S^{1,k}_n(p)
$$ 
and
$$
S^2_n(p)=\sum\limits_{k=1}^\infty\epsilon_k S^{2,k}_n(p).
$$
It can be proved by the mathematical induction on $n$ 
that $0\le Q_n^{i,k}\le 2^n$ and $|S^{i,k}_n(p)|\le 2^{n-1}$
for $i=1,2$ and for all $k$, $p$ and $n$.
Then these sums are finite for each $n$ and $p$.

By (\ref{cap-1t})-(\ref{cap-2t}) the Skeptic's capital ${\cal Q}_n$ 
at step $n$, when he follows the strategy $S_n(p)$, equals                                      
$$
{\cal Q}_n=\frac{1}{2}\sum\limits_{k=1}^\infty
\epsilon_k ({\cal Q}^{1,k}_n+{\cal Q}^{2,k}_n).
$$
Define for each $n$ the function
$g_n(p)=(2\xi(p\le 0.5)-1)(\omega_n-p)$. Let
$E_{P_n}(g_n)=\int g_n(p)P_n(dp)$.

Recall that Forecaster uses some randomized strategy $P_n$, $n=1,2,\dots$.

We define recursively by $n$: ${\cal F}^{k}_0=1$, $g^k_0(p)=0$, 
and for $n\ge 1$
\begin{eqnarray}
g^k_n(p)=-\epsilon_k{\cal F}^k_{n-1}(g_n(p)-E_{P_n}(g_n)),
\label{strat-1ran}
\end{eqnarray}
where $\epsilon_k=2^{-k}$, and for $n\ge 1$
\begin{eqnarray}
{\cal F}^k_{n}={\cal F}^k_{n-1}+g^k_{n}(p_{n})
\label{strat-1rani}
\end{eqnarray}
By definition for any $k$ and $n$,
\begin{eqnarray}
{\cal F}^k_n=\prod\limits_{j=1}^n (1-\epsilon_k(g_j(p_j)-E_{P_j}(g_j))).
\label{cap-1ran}
\end{eqnarray}
By (\ref{cap-1ran}) $0\le {\cal F}^k_n\le 2^n$ for all $n$ and $k$. 

Finally, Skeptic defines at step $n$
$$
f_n(p)=\sum\limits_{k=1}^\infty\epsilon_k g^k_n(p).
$$
By definition $\int f_n(p)P_n(dp)\le 0$.

By (\ref{cap-1ran}) the Skeptic's capital ${\cal F}_n$ at step $n$, 
when he follows the strategy $f_n(p)$, equals
$$
{\cal F}_n=\sum\limits_{k=1}^\infty\epsilon_k {\cal F}^k_n.
$$
Also, ${\cal F}_n\ge 0$ for all $n$.

Suppose that $\sup\limits_{n}{\cal F}_n=C<\infty$, where $C>0$.
Then $\sup\limits_{n}{\cal F}^k_n<\frac{C}{\epsilon_k}$ for each $k$.

We have for each $k$,
\begin{eqnarray*}
\ln {\cal F}^k_n\ge-\epsilon_k\sum\limits_{j=1}^n
(g_j(p_j)-E_{P_j}(g_j))-n\epsilon_k^2.
\end{eqnarray*}
Here we use the inequality
$\ln (1+r)\ge r-r^2$ for all $|r|\le\frac{1}{2}$.

Since ${\cal F}_n$ is bounded by $C>0$, we have for any $k$
\begin{eqnarray}
\frac{1}{n}\sum\limits_{j=1}^n (g_j(p_j)-E_{P_j}(g_j))\ge
\frac{-\ln C+\ln(\epsilon_k)}{n\epsilon_k}-
\epsilon_k\ge-2\epsilon_k
\label{inneq-1}
\end{eqnarray}
for all sufficiently large $n$.

Define two variables
\begin{eqnarray*}
\vartheta_{n,1}=\sum\limits_{j=1}^n\xi(p_j>0.5)(\omega_j-p_j),
\label{var-1a}
\\
\vartheta_{n,2}=\sum\limits_{j=1}^n\xi(p_j\le 0.5)(\omega_j-p_j).
\label{var-2a}
\end{eqnarray*}

By definition of $g_j$,
$$
\vartheta_{n,2}-\vartheta_{n,1}=\sum\limits_{j=1}^n g_j(p_j).
$$
For technical reason define $g_{1,j}(p)=\xi(p>0.5)(\omega_j-p)$ and
$g_{2,j}(p)=\xi(p\le 0.5)(\omega_j-p)$. Then $g_j(p)=g_{2,j}(p)-g_{1,j}(p)$.

Assume for any $n$ the probability distribution $P_n$ is concentrated on
a finite set $\{p_{n,1},\dots ,p_{n,m_n}\}$.

For technical reason, if necessary, we add $0$ and $1$ to the
support set of $P_n$ and set their probabilities to be $0$.
Denote $p_n^-=\max\{p_{n,t}:p_{n,t}\le 0.5\}$ and
$p_n^+=\min\{p_{n,t}:p_{n,t}>0.5\}$.

By definition $\omega_n$, $p_n^+$ and
$p_n^-$ are predictable and $p_n^+-p_n^-\ge\Delta$ for all $n$,
where $\Delta>0$. We have
\begin{eqnarray}
\sum\limits_{j=1}^n E_{P_j}(g_{1,j})\le
\sum\limits_{\omega_j=0} P_j\{p>0.5\}(-p_j^+)+
\sum\limits_{\omega_j=1} P_j\{p>0.5\}(1-p_j^+)\le
\nonumber
\\
-0.5\sum\limits_{j=1}^n\xi(\omega_j=0)p_j^+
+0.5\sum\limits_{j=1}^n\xi(\omega_j=1)(1-p_j^+).
\label{expect-1a}
\\
\sum\limits_{j=1}^n E_{P_j}(g_{2,j})\ge
\sum\limits_{\omega_j=0} P_j\{p\le 0.5\}(-p_j^-)+
\sum\limits_{\omega_j=1} P_j\{p\le 0.5\}(1-p_j^-)\ge
\nonumber
\\
-0.5\sum\limits_{j=1}^n\xi(\omega_j=0)p_j^-+
0.5\sum\limits_{j=1}^n\xi(\omega_j=1)(1-p_j^-).
\label{expect-2a}
\end{eqnarray}

Subtracting (\ref{expect-1a}) from (\ref{expect-2a}), we obtain
$$
\sum\limits_{j=1}^n E_{P_j}(g_j)=\sum\limits_{j=1}^n E_{P_j}(g_{2,j})-
\sum\limits_{j=1}^n E_{P_j}(g_{1,j})\ge 0.5\Delta n.
$$

Using (\ref{inneq-1}), we obtain for all sufficiently large $n$
\begin{eqnarray}
\frac{1}{n}(\vartheta_{n,2}-\vartheta_{n,1})=
\frac{1}{n}\sum\limits_{j=1}^n g_j(p_j)\ge\frac{1}{n}
\sum\limits_{j=1}^n E_{P_j}(g_j)-2\epsilon_k\ge 0.5\Delta-2\epsilon_k.
\label{oo-1}
\end{eqnarray}

Now we compute a lower bound of Skeptic's capital. 

We have from the definition (\ref{strat-1})-(\ref{strat-2}) and 
(\ref{cap-1})-(\ref{cap-2}).

\begin{eqnarray}
{\cal Q}_n^{1,k}=\prod\limits_{j=1}^n (1-\epsilon_k\xi(p_j>0.5)(\omega_j-p_j)),
\label{cap-1}
\\
{\cal Q}_n^{2.k}=\prod\limits_{j=1}^n (1+\epsilon_k\xi(p_j\le 0.5)
(\omega_j-p_j)).
\label{cap-2}
\end{eqnarray}
By (\ref{cap-1}) and (\ref{cap-2}), for $i=1,2$, 
$0\le {\cal Q}_n^{i,k}\le 2^n$ for all $n$
no matter how the other players move.

By~(\ref{cap-1})-(\ref{cap-2}) at step $n$
\begin{eqnarray}
\ln {\cal Q}_n^{1,k}\ge -\epsilon_k \vartheta_{n,1}-\epsilon_k^2n,
\label{cap-1a}
\\
\ln {\cal Q}_n^{2,k}\ge\epsilon_k\vartheta_{n,2}-\epsilon_k^2n.
\label{cap-2a}
\end{eqnarray}

The inequalities (\ref{cap-1a}), (\ref{cap-2a}) and (\ref{oo-1}) imply
\begin{eqnarray}
\limsup\limits_{n\to\infty}
\frac{\ln {\cal Q}_n^1+\ln {\cal Q}_n^2}{n}\ge
0.5\epsilon_k\Delta-2\epsilon_k^2
\ge 2\epsilon_k^2
\label{exp-3a}
\end{eqnarray}
for all sufficiently large $n$, where $\epsilon_k\le\frac{1}{8}\Delta$.

From this, we obtain that for $i=1$ or for $i=2$, 
\begin{eqnarray*}
\limsup\limits_{n\to\infty}
\frac{\ln {\cal Q}_n^{i,k}}{n}\ge\epsilon_k^2
\label{exp-3aa}
\end{eqnarray*}
for all sufficiently large $n$.

Hence, we obtain for the total capital of Skeptic 
${\cal K}^n={\cal Q}^n+{\cal F}^n$  
$$
\limsup\limits_{n\to\infty}{\cal K}_n=\infty
$$
no matter how Forecaster moves if Realty uses her strategy defined above.

We obtain also a lower bound of calibration error for Binary Forecasting Game II.
\begin{corollary}
Assume Forecaster's uses a randomized strategy with
a positive level of discreteness on each infinite sequence $\omega$.
Then Realty can announce an infinite binary sequence $\omega_1\omega_2\dots$
such that
\begin{eqnarray}
\limsup\limits_{n\to\infty}\left|\frac{1}{n}\vartheta_{n,i}\right|\ge
0.25\Delta
\label{lower-bound}
\end{eqnarray}
for $i=1$ or for $i=2$.
\end{corollary}
This inequality immediately follows from (\ref{oo-1}).

\end{document}